\documentclass{article}
\usepackage[preprint]{neurips_2026}
\usepackage{longtable}
\usepackage{booktabs}
\usepackage{array}
\usepackage[table]{xcolor}
 \usepackage{booktabs}
 \usepackage{tabularx}
 \usepackage{xcolor}
\usepackage{caption}
\newcommand{\wc}[3][{}]{{C_{\! \underset{#3}{#2}}^{{#1}}} }

\usepackage[utf8]{inputenc}
\usepackage[T1]{fontenc}
\usepackage{hyperref}
\usepackage{url}
\usepackage{booktabs}
\usepackage{amsfonts}
\usepackage{amsmath}
\usepackage{amssymb}
\usepackage{bm}
\usepackage{nicefrac}
\usepackage{microtype}
\usepackage{xcolor}
\usepackage{graphicx}
\usepackage{float}
\usepackage{orcidlink}

\title{ A Reinforcement Learning Method to Navigate SMEFT Operator Space}

\author{%
  Jacky Kumar\,\orcidlink{0000-0001-9053-0731} \\ 
  Physique des Particules \\
  Universit\'e de Montr\'eal \\
  Montr\'eal, QC H2V 0B3, Canada \\
  \texttt{jacky.kumar@umontreal.ca}
}

\begin{document}
\maketitle
\begin{abstract}

In particle physics, discrepancies between experimental measurements and Standard Model (SM) predictions are commonly interpreted within  new theoretical models. In this context, the Standard Model Effective Field Theory (SMEFT) provides a general framework encompassing a broad class of models. However, exploring the enormous space of possible SMEFT operators rapidly becomes a challenging combinatorial problem: many operators can contribute simultaneously to different observables, their effects are correlated, and distinct operator combinations can provide comparably good descriptions of the data.
In this work, we formulate the exploration of the SMEFT operator space as a sequential decision-making problem using reinforcement learning (RL).
Our RL agent employs a lightweight ``actor--critic" architecture with approximately 40K trainable parameters in total.
Rather than searching for a single best-fit theory, our method is designed to discover a diverse set of high-quality theory hypotheses. We test the method in a complex phenomenological environment comprising 132 observables across several sectors of particle physics, including 24 measurements that deviate from their SM predictions by more than $1.5\sigma$. We show that the learned RL policy efficiently constructs multi-operator theories that substantially improve the global fit. Comparisons with independent random-subset and random-sequential searches demonstrate the ability of the RL policy to navigate this large combinatorial space. Our results establish RL as a systematic framework for exploring diverse patterns of new physics (NP) interactions within SMEFT.\footnote{ The implementation of our RL framework and the accompanying analysis tools and data are publicly available as open-source software: \href{https://github.com/jackypheno/ReinforceSMEFT}{ReinforceSMEFT}.}
\end{abstract}

%\newpage

\section{Introduction}
\label{sec:introduction}

The SM provides a successful
description of particle physics over a wide range of energy scales.
Nevertheless, it cannot account for several important observations such as
neutrino masses, dark matter, and the baryon asymmetry of the
universe. This motivates the presence of physics beyond the SM (BSM). At the same time, direct searches at the Large Hadron
Collider (LHC) have so far revealed no conclusive evidence for such NP. 
If the scale of NP lies much above the energies
directly accessible to current colliders, its effects may instead first
appear indirectly through deviations of low-energy observables from their
SM predictions, i.e., anomalies.

Effective field theories (EFTs) provide a model-independent framework
for interpreting such discrepancies. Above the electroweak scale, the
effects of heavy NP can be parameterized within the Standard Model
Effective Field Theory (SMEFT)
\cite{Buchmuller:1985jz,Grzadkowski:2010es}(for the reviews, see \cite{Brivio:2017vri,Isidori:2023pyp, Aebischer:2025qhh}),
which contains only SM fields and respects the 
$SU(3)_C\times SU(2)_L\times U(1)_Y$ gauge symmetry. Below the
electroweak scale, the appropriate description is the Weak Effective
Theory (WET), obtained after integrating out the heavy SM degrees of
freedom and respecting $SU(3)_C\times U(1)_{\rm em}$.

\paragraph{Statement of the problem:}
A central problem in indirect searches for NP is the following:
given a collection of experimental measurements, which combinations of
SMEFT operators can improve the agreement with the data while remaining
compatible with the other experimental constraints? This bottom-up
problem is an important intermediate step toward the eventual top-down
question of identifying UV theories capable of generating the preferred
SMEFT operators.

However, the difficulty arises from the enormous combinatorial space of the
SMEFT operators. At dimension six, the Warsaw basis \cite{Grzadkowski:2010es} contains 59
baryon and lepton number conserving operators, corresponding to 2499 independent operators
once flavour indices are taken into account.
Even when only a subset of these operators is relevant to a particular
low-energy dataset, the number of possible multi-operator hypotheses grows rapidly.
For $N$ candidate operators, the number of possible theories containing
up to $k$ operators is
\begin{equation}
N_{\rm comb}
=
\sum_{n=1 }^{k}
{N\choose n}.
\end{equation}
Here $N$ is total number of SMEFT operators considered in a problem, and $n$ represents the 
number of operators in a given SMEFT hypothesis. In the
present analysis, we work with 912 SMEFT operators that conserve the 
baryon number, lepton number, and lepton flavour. 
The challenge is therefore not simply to determine the continuous values
of a set of WCs, but first to determine
\emph{which operators should be present}. This introduces a discrete
combinatorial search problem on top of the usual continuous parameter
estimation. 

Importantly, SMEFT operators cannot generally be considered independent because
a single operator may affect multiple observables. These correlations become
even richer after renormalization-group (RG) evolution and matching
between SMEFT and WET \cite{Aebischer:2025qhh}. Operator mixing due to RG running 
implies that an operator introduced to improve one observable may simultaneously modify several
others. In order to interpret experimental measurements within SMEFT, therefore it requires  
understanding of non-trivial correlations among different directions in the operator space.

These correlations are particularly important for multi-operator
SMEFT hypotheses. An operator that improves the description of one discrepant
observable may worsen agreement with another measurement. A second
operator can then compensate for this deterioration while preserving, or
further improving, the original effect. Subsequent operators can
similarly repair constraints or address additional discrepancies.
Consequently, the quality of a multi-operator hypothesis cannot in general
be inferred from the performance of its constituent operators in
isolation. Viable solutions can instead emerge through correlated
combination of operators.

Conventional phenomenological analyses address this complexity using
physics intuition to restrict the set of operators under consideration,
often guided by a particular UV theory, symmetry argument, or simplified
model. Such approaches are very useful for testing specific
hypotheses, but necessarily explore only a small fraction of the complete
SMEFT operator space. At the opposite extreme, exhaustive enumeration of
all possible operator combinations quickly becomes computationally
prohibitive. In this work, we formulate the search for viable SMEFT  
operators a sequential decision making 
problem using an ``actor-critic" RL algorithm \cite{sutton1999policy, sutton2018reinforcement}
 as introduced below.

\paragraph{Reinforcement learning method:}
We design an RL policy using a transformer \cite{vaswani2023attentionneed} based neural-network (NN) which learns to propose 
 viable SMEFT operators combinations through repeated interactions with phenomenological environment. So instead of evaluating
every SMEFT operator combination independently, an RL policy  can construct
a multi-operator hypothesis sequentially, observe how the global fit to experimental 
data changes, and use this feedback to 
adjusts the NN weights to learn. 

An important feature of this formulation is that the policy can learn
\emph{conditional} operator preferences. If the addition of one operator
repeatedly produces a characteristic pattern of improvements and
deterioration among observables, the policy can learn which subsequent
operators are effective in that particular state. The RL agent \footnote{We use the terms RL agent and RL policy interchangeably throughout the text.} therefore
does not merely learn a global ranking of individual operators, it learns
how useful an operator is conditional on the presence of other operators.

We stress that our objective is broader than identifying a single best-fit theory.
Different combinations of SMEFT operators can provide comparably good
descriptions of the same data, and identifying this degeneracy is itself
 important for model building. We therefore formulate the task as a
\emph{quality--diversity search}: the agent should discover high-quality
solutions while continuing to explore structurally different regions of
the SMEFT operator space. To achieve this, we supplement the fit-quality
reward \cite{Kumar:2026zpo} with a novelty-based reward function that discourages repeated sampling of
closely related operator sets. The details of RL formulation are given in the next section. 

Related efforts have recently explored the use of machine learning (ML) and artificial intelligence (AI) techniques for automated theory exploration in particle physics. For example, Ref.~\cite{Baretz:2025zsv} developed the Autonomous Model Builder (AMBer), which employs RL to construct viable neutrino-flavor models by selecting symmetry groups, particle content, and representation assignments. Ref.~\cite{Alexander:2026lpw} proposed the autonomous discovery of particle physics theories directly from experimental data using RL and symbolic-regression techniques. Ref.~\cite{Hammad:2026bvw} explored a complementary language-model-based approach in which a domain-adapted LLM generates hypotheses for sparse SMEFT analyses, while Ref.~\cite{Saad:2026pan} developed an LLM-assisted framework for BSM model building, in which an LLM provides a natural-language interface to a deterministic symbolic backend that automatically constructs and analyzes BSM models from user-specified field content.

These studies illustrate the growing interest in using ML to navigate large theory space and facilitate traditionally human-guided aspects of model building in particle physics. 

\paragraph{Code availability:}
The method presented in this work builds upon my open-source GitHub implementation \href{https://github.com/jackypheno/ReinforcePhysics}{ReinforcePhysics}. This RL framework includes implementations based on both the ``actor--critic" algorithm and Group Relative Policy Optimization (GRPO)\cite{GRPO}; in the present work, we focus on the former. To facilitate reproducibility and broader use, we have packaged the complete methodology into the pip-installable Python package, which is publicly available and will be actively maintained in the \href{https://github.com/jackypheno/ReinforceSMEFT}{ReinforceSMEFT} GitHub repository.

\section{Reinforcement Learning Formulation}
\label{sec:method}
In what follows, we provide details of our RL formulation of the SMEFT problem and its implementation.
 Starting from the SM, the RL agent
proposes one SMEFT operator at each time step $t$. The Wilson coefficients (WCs) of the
currently proposed operators are optimized by minimizing the global
$\chi^2$, and the resulting improvement in the fit along with diversity criteria provides the primary
feedback to the agent in terms of rewards. The correspondence between the RL and the SMEFT problem is, therefore:
\begin{itemize}
    \item \textbf{Action} $a_t$: proposing an SMEFT operator;
    \item \textbf{Reward} $r_t$: improvement in fit quality  and solution diversity\footnote{Note that in this method the reward function can be easily modified and defined according to specific phenomenological preferences.   };
    \item \textbf{State} $s_t$: contains the features -- current observable pulls, $\chi^2$, selected
    operators, previously evaluated operators, and trajectory step $t$;
      \item \textbf{Policy} $\pi_\theta(a_t|s_t)$:  is the rule of decision to be learned that assigns
    probabilities to candidate SMEFT operators or actions $a_t$, $\theta$ are learnable parameters;
     \item \textbf{Environment:} observable predictions within the SMEFT,
    input experimental measurements, global $\chi^2$ minimizer. 
\end{itemize}
The pull of the $i$th observable is defined as
$p_i=(O_i^{\rm exp}-O_i^{\rm SM})/\sqrt{(\sigma_i^{\rm exp})^2+(\sigma_i^{\rm SM})^2}$,
where $O_i^{\rm exp}$ and $O_i^{\rm SM}$ denote the experimental measurement and the SM prediction, respectively, while $\sigma_i^{\rm exp}$ and $\sigma_i^{\rm SM}$ are their corresponding uncertainties. A positive (negative) pull indicates that the experimental value lies above (below) the SM prediction.

The policy sequentially selects SMEFT operators, thereby generating a trajectory of states
\begin{equation}
s_0
\xrightarrow{\;O_1\;}
s_1
\xrightarrow{\;O_2\;}
s_2
\xrightarrow{\;O_3\;}
\cdots
\xrightarrow{\;O_T\;}
s_T 
\label{eq:rl_trajectory}
\end{equation}
where $O_t$ denotes the operator selected at step $t$, and $s_t$ represents the state of the environment after the first $t$ operators have been selected. 
Initially, a proposed operator is temporarily added to the currently accepted
operator set and the corresponding $\chi^2$ is minimized. The operator is
retained if it improves the fit by more than $\Delta\chi^2_{\rm min}=2$ 
and is rejected otherwise. In either case, it is marked as evaluated so
that the same operator is not proposed again during the trajectory.
The process is repeated for at most $T=10$ steps (i.e., the trajectory length).

At each step, the state encodes information about the current quality of the global fit, together with the operators selected and evaluated up to that point. Based on the current state, the policy assigns a probability to each available operator
\begin{equation}
O_{t+1}\sim\pi_\theta(O\,|\,s_t),
\end{equation}
and the selected operator determines the transition from $s_t$ to $s_{t+1}$. During training, the policy learns to assign larger probabilities to operator selections that lead to larger cumulative future rewards. Consequently, the agent is not restricted to selecting operators that provide the largest immediate improvement in the fit; it can learn sequences in which an operator produces only a modest initial improvement but enables subsequent operators to achieve a substantially better global fit.

During the RL training itself we save all possible operator hypotheses (here a single hypothesis is defined by accepted operators in a trajectory in $T$ time steps) sampled from the policy and post-training we recover all 
viable hypotheses based on some criteria; such as minimum $\Delta \chi^2$ or number of operators.  This approach is similar to Deep Symbolic Regression \cite{DSR} which 
discovers mathematical expressions describing a dataset during the RL training.

\subsection{Actor--critic learning}
\label{sec:actorcritic}

The ``actor--critic" algorithm is used to train the policy. The basic idea of the 
actor--critic algorithm is straightforward:
actor is simply the policy $\pi_\theta(a_t|s_t)$, which we parameterize by a small transformer-based NN. 
It determines the probability of selecting action $a_t$ (or an SMEFT operator $O_t$)
 in a given state $s_t$. The critic estimates the expected future return from the current
state and thereby provides a baseline for the policy-gradient update.

Here, the return is defined by the cumulative discounted reward obtained from the current state onward when following the current policy $\pi_\theta$. For a sampled trajectory, the return at step $t$ is
$G_t=\sum_{k=0}^{T-t}\gamma^k r_{t+k}$, where $r_{t+k}$ is the reward received at a future 
step and $\gamma$ is the discount factor.

The actor, with learnable parameters $\theta$, is optimized using the loss
\begin{equation}
\mathcal{L}_{\rm actor}(\theta)
=
-\frac{1}{N}
\sum_t
A_t
\log\pi_\theta(a_t|s_t).
\label{eq:actorloss}
\end{equation}
Here, $A_t$ is the advantage, which quantifies whether the
sampled action performs better or worse than the expected behaviour of the
policy from that state. In terms of the so-called ``true value function" $V^\pi(s_t)$,
the advantage is defined as
\begin{equation}
A_t
=
G_t-V^\pi(s_t).
\end{equation}
However, since the true value function $V^\pi(s_t)$ is not known, we train a separate
NN, the critic, to approximate it by $V_\phi(s_t)$, where $\phi$
denotes the learnable critic parameters. 

For each batch of sampled trajectories, the
critic is trained using the objective
\begin{equation}
\mathcal{L}_{\rm critic}(\phi)
=
\frac{1}{N}
\sum_t
\left[V_\phi(s_t)-G_t\right]^2.
\end{equation}
Here, $N$ denotes
the total number of sampled state--action steps entering the corresponding
training update.
The estimated value function $V_\phi(s_t)$ is then used to construct the advantage:
$A_t \approx G_t-V_\phi(s_t)$,
which enters the actor loss in Eq.~\eqref{eq:actorloss}. 

The total training objective combines the actor and critic losses
\begin{equation}
\label{eq:combinedloss}
\mathcal{L}_{\rm total}
=
\mathcal{L}_{\rm actor}
+
\lambda_V\mathcal{L}_{\rm critic}
\end{equation}
where $\lambda_V=0.5$ controls the relative contribution of the critic loss.

Training proceeds iteratively by sampling batches of trajectories from the
current policy $\pi_\theta$ and using them to update the network parameters.
For the actor, the gradient $\nabla_\theta\mathcal{L}_{\rm actor}$ increases
the probability of actions with positive advantage and suppresses those with
negative advantage. Importantly, $A_t$ is treated as a fixed target when
computing the actor gradient. Consequently, for separate actor and critic
networks, the critic loss does not contribute directly to the gradient with
respect to $\theta$; instead, the critic influences the actor update through
the state-dependent baseline $V_\phi(s_t)$ used to construct the advantage.

In our implementation, $\mathcal{L}_{\rm actor}$ and
$\mathcal{L}_{\rm critic}$ are combined into the single objective in
Eq.~\eqref{eq:combinedloss} and optimized in one backward pass. An RL
algorithm that does not employ a critic and instead uses Monte Carlo returns
directly in the policy-gradient estimator is commonly known as REINFORCE
\cite{williams1992simple}. The principal role of the critic is therefore to
reduce the variance of the policy-gradient estimate by providing a
state-dependent baseline, while the actor learns which operator selections
lead to larger-than-expected returns.

\subsection{State representation and NN architecture}
\label{sec:rl_architecture}

At each step $t$, the RL agent observes the Markov state
\begin{equation}
    s_t =
    \left(
    \mathbf{p}_t,\,
    \chi_t^2,\,
    \mathbf{m}^{\rm active}_t,\,
    \mathbf{m}^{\rm eval}_t,\,
    t
    \right)
\end{equation}
where $\mathbf{p}_t=(p_{1,t},\ldots,p_{N_{\rm obs},t})$ contains the pulls which quantify the deviations of the observables w.r.t. the SM, $\chi_t^2$ quantifies the overall agreement with the data, $\mathbf{m}^{\rm active}_t$ specifies the operators currently included in the SMEFT scenario, and $\mathbf{m}^{\rm eval}_t$ records the operators that have already been evaluated during the trajectory. The latter prevents the duplicate operator from being proposed repeatedly. The step number $t$ provides information about the position within the trajectory. Both the actor and critic receive this same state representation.

As we have a fixed number of SMEFT operators in the problem,  
each candidate operator $O_i$ can be represented by a trainable embedding
\begin{equation}
    O_i \longrightarrow
    \mathbf{e}^{\,\rm op}_i \in \mathbb{R}^{d}.
\end{equation}
In our implementation, the latent dimension ($d$) is 32. 
Each observable pull is projected independently into the same latent space 
\begin{equation}
    p_{j,t}\longrightarrow
    \mathbf{e}^{\,p}_{j,t}
    =W_p p_{j,t}+\mathbf{b}_p
    \in\mathbb{R}^{d}.
\end{equation}
Here, $W$ and $\mathbf{b}$ denote a subset of the trainable parameters of the NN. 
To relate the candidate operators to the current pattern of deviations in the observable space, we employ multi-head cross-attention, taking the operator embedding as queries and the pull embedding as keys and values, i.e.
$Q=E^{\rm op},
    K=E^p(s_t)$ and $ V=E^p(s_t)$.
    
For a single attention head, this operation takes the standard form \cite{vaswani2023attentionneed}
\begin{equation}
    {\rm Attn}(Q,K,V)
    =
    {\rm softmax}
    \left(
        \frac{QK^{T}}{\sqrt{d_h}}
    \right)V
\end{equation}
here, we use four attention heads, i.e., $h=4$. 

The resulting vector
$\mathbf{c}_{i,t}\in\mathbb{R}^{d}$ -- the $i$-th row of Attn -- provides a state-dependent representation of the observable information relevant to candidate operator $O_i$. In particular, the attention mechanism allows different operators to focus on different patterns of observable pulls.

The remaining components of the state are encoded separately. The binary active and evaluated operators are mapped to trainable embedding
\begin{equation}
    m^{\rm active}_{i,t}
    \longrightarrow
    \mathbf{e}^{\,\rm active}_{i,t},
    \qquad
    m^{\rm eval}_{i,t}
    \longrightarrow
    \mathbf{e}^{\,\rm eval}_{i,t}
\end{equation}
while the normalized global quantities
$(
    {\chi_t^2} / {\chi_{\rm ref}^2},
    {t}/{T_{\max}}
)
$
are passed through a small multi-layer perceptron to obtain a global representation
$\mathbf{g}_t\in\mathbb{R}^{d}$. 

The complete state-dependent representation of operator $O_i$ is then constructed by concatenating these components
\begin{equation}
\boxed{
    \mathbf{h}_{i,t}
    =
    \left[
    \mathbf{e}^{\,\rm op}_i;\,
    \mathbf{c}_{i,t};\,
    \mathbf{e}^{\,\rm active}_{i,t};\,
    \mathbf{e}^{\,\rm eval}_{i,t};\,
    \mathbf{g}_t
    \right]
    \in\mathbb{R}^{5d}.
}
\label{eq:operator_state}
\end{equation}
For $d=32$, each candidate operator is represented by a $160$-dim vector. This representation encodes the operator identity together with its attention-weighted relation to the current observable pulls, whether it has already been selected or evaluated, and information about the current global state of the trajectory.
From a physics perspective, this representation allows the relevance of each operator to depend dynamically on the pattern of deviations of observables from the SM and on the operators already selected, thereby enabling the policy to learn correlations and complementary effects among different operators.

The actor maps each representation $\mathbf{h}_{i,t}$ to a scalar logit $z_{i,t}$
\begin{equation}
    z_{i,t}=f_{\theta}(\mathbf{h}_{i,t}),
\end{equation}
from which the policy is obtained as
\begin{equation}
    \pi_{\theta}(O_i|s_t)
    =
    \frac{\exp(z_{i,t})}
    {\displaystyle\sum_{j\in\mathcal{A}_t}
     \exp(z_{j,t})},
\end{equation}
where $\mathcal{A}_t$ denotes the set of operators that have not yet been evaluated. Previously evaluated operators are explicitly masked and hence have zero probability of being selected again. 

As the role of critic is to determine how promising the current state is, irrespective of which operator we choose, it only needs one global representation of the current state, which is obtained using 
\begin{equation}
    \overline{\mathbf{h}}_t
    =
    \frac{1}{N_{\rm op}}
    \sum_{i=1}^{N_{\rm op}}
    \mathbf{h}_{i,t}.
\end{equation}
Here ${N_{\rm op}}$ is number of sampled operators at time step $t$.
After concatenating with the global quantities $\chi^2_t,t$, i.e., $\mathbf{g_t}$ the critic then 
predicts the scalar state value
\begin{equation}
    V_{\phi}(s_t)
    =
    f_{\phi}
    \left(
    [\,\overline{\mathbf{h}}_t;\mathbf{g}_t\,]
    \right).
\end{equation}
Thus, the actor learns which operator should be selected next, whereas the critic estimates the expected quality of the current state. 

Crucially, the operator embedding, pull projection, attention mechanism, state embedding, actor, and critic are trained jointly through the RL objective. Consequently, the network learns a \emph{state-dependent} representation of the SMEFT operator space: the preference for a given operator can change as the pattern of pulls and the set of previously selected operators evolve along a trajectory.

The NN architecture and RL training framework are implemented using \textsc{PyTorch}~\cite{paszke2019pytorch}.
For the full action space of 912 SMEFT operators and an embedding dimension $d=32$, the actor-critic NNs contain approximately  $\sim$40K trainable parameters and is sufficiently lightweight to be trained efficiently using standard CPU resources, without requiring GPU acceleration. In practice, the computational cost is dominated by repeated evaluation and optimization of the physics environment (which is mainly the global $\chi^2$-minimization) rather than by the NN operations. 

\subsection{Reward function}
In our setup, at each interaction, the environment provides feedback to the policy through a scalar reward signal. In addition to improving the quality of the fit, our goal is to encourage the policy to explore distinct regions of the SMEFT operator space. Accordingly, the reward function consists of two components: a physics reward associated with the improvement in fit quality and a diversity reward that promotes the discovery of distinct operator combinations. The diversity reward is evaluated and added at the end of each trajectory. The total reward is therefore defined as
\begin{align}
r_t
&=
r_t^{\rm fit}
+
\lambda_{\rm div}
\left(1-J_{\max}^{\mathcal A}\right)\, \mathbb{I}(t=T).
\label{eq:reward}
\end{align}
 Here, parameter $\lambda_{\rm div}$ controls the relative
importance of diversity compared to improvement of the physics fit and 
 $\mathbb{I}(t=T)$ is an indicator function that equals $1$ only at the final step $t=T$, and $0$ at all earlier steps, ensuring that the novelty reward is added only at the end of the trajectory. 

The fit reward, first term of Eq.~\ref{eq:reward}, is proportional to the improvements in the fit, given by 
\begin{align} \label{eq:rewardfit}
    r_t^{\rm fit}
&=
\Delta\chi_t^2
-\alpha_{\rm step}
-\lambda_{\rm dead}\,
\mathbb{I}\!\left(\Delta\chi_t^2 < \Delta\chi_{\rm min}^2\right)
+\lambda_{\rm sol}\,
\mathbb{I}\!\left(\chi_t^2 < \chi_{\rm sol}^2\right).
\end{align}
Here, $\Delta\chi_t^2=\chi_{t-1}^2-\chi_t^2$ denotes the reduction in the global
$\chi^2$ obtained after the action at step $t$, such that a positive
value corresponds to an improvement of the fit.
The parameter $\alpha_{\rm step}$ provides a small cost for each
operator-selection step, favoring solutions with small a number of operators.
The third term penalizes unproductive actions, with
$\lambda_{\rm dead}$ denoting the dead-step penalty and
$\Delta\chi_{\rm min}^2$ the minimum improvement required for an
action to be considered useful. The final term in
 Eq.~\ref{eq:rewardfit} provides an additional reward
$\lambda_{\rm sol}$ when the resulting theory reaches the target
fit quality $\chi_{\rm sol}^2$.

For the diversity reward given by the second term of Eq.~\ref{eq:reward}, the similarity between two
operator sets is quantified by the Jaccard index: $J(S_T,S')={|S_T\cap S'|}/{|S_T\cup S'|}$.  
The largest similarity of current trajectory $S_T$ to any other set already contained in the archive $\mathcal A$ is given by
$J_{\max}^{\mathcal A}=\max_{S'\in\mathcal A} J(S_T,S')$, here,  $\mathcal A$ is an archive of previously
discovered high-quality solutions.  Only solutions satisfying the chosen quality requirement,
$\Delta\chi^2>\Delta\chi^2_{\rm archive}$, are added to this diversity archive.
Consequently, a solution identical or very similar to an archived
solution receives little diversity reward, whereas a solution occupying
a previously unexplored region of operator space receives a larger
reward.

\section{RL Environment}
\label{sec:environment}

The SMEFT operators are sampled sequentially using the current policy $\pi_\theta$. 
After each operator is sampled in a trajectory, our framework invokes an internal $\chi^2$-minimization procedure to evaluate the resulting fit and determine the corresponding reward. Since training the RL policy requires a large number of such interactions with the environment, this procedure must be repeated many times. A direct evaluation of the $\chi^2$ function would require performing the full RG evolution within SMEFT and WET, followed by the computation of the relevant observables, for every proposed set of operators. Combined with the repeated $\chi^2$ minimization, this would render the RL training prohibitively expensive.

To overcome this computational bottleneck, we precompute semi-analytic expressions for the observables, retaining their linear dependence on the SMEFT WCs defined at the NP scale. The computationally expensive RG evolution and observable calculations are therefore performed only once during this preprocessing stage. The resulting expressions can subsequently be evaluated very efficiently within the RL environment, enabling rapid computation of the $\chi^2$ function and the corresponding rewards throughout training.

\subsection{Semianalytic $\chi^2$ function}
\label{sec:semi_analytic}

At dimension-six level in SMEFT, the theoretical prediction for an observable $a$ can be written as
\begin{equation}
\mathcal{O}_{{\rm th},a}(\{C_i\})
=
\mathcal{O}_{{\rm SM},a}
+
\sum_i \alpha_{ai}C_i(\Lambda)
+
\sum_{j,k,\, j \le k}\beta_{a,jk}
C_j(\Lambda)C_k(\Lambda)
\label{eq:semi_analytic_obs}
\end{equation}
where $C_i(\Lambda)$ are the WCs defined at the NP scale $\Lambda$. The coefficients $\alpha_{ai}$ and $\beta_{a,jk}$ encode, respectively, the linear and quadratic dependence of observable $a$ on the SMEFT WCs, including the effects of RG evolution and matching at the electroweak scale. We assume a CP-conserving SMEFT scenario, such that the WCs are real. In the present analysis, we retain only the linear contribution proportional to $\alpha_{ai}$ and construct the corresponding semi-analytic expression for each observable included in the global fit. We emphasize, however, that the RL method itself is not restricted to the linear approximation and can also be applied when quadratic terms are retained.

The $\chi^2$ function is defined as
\begin{equation}
\chi^2(\{C_i\})
=
\sum_a
\frac{
\left[
\mathcal{O}_{{\rm th},a}(\{C_i\})
-
\mathcal{O}_{{\rm exp},a}
\right]^2
}{
\sigma_{{\rm th},a}^2+
\sigma_{{\rm exp},a}^2
}
\label{eq:chi2_smeft}
\end{equation}
where $\mathcal{O}_{{\rm exp},a}$ denotes the experimental measurement, while $\sigma_{{\rm th},a}$ and $\sigma_{{\rm exp},a}$ denote the corresponding theoretical and experimental uncertainties, respectively. To determine the semi-analytic coefficients, we use \texttt{flavio}~\cite{Straub:2018kue} for the computation of observable predictions and \texttt{wilson}~\cite{Aebischer:2018bkb} for RG evolution.

The computationally expensive SMEFT calculations are therefore performed only once when constructing the semi-analytic expressions, while the subsequent $\chi^2$ evaluations required repeatedly during RL training can be carried out efficiently using these precomputed expressions. We find that the RL training would be extremely slow without following this procedure.

\subsection{Benchmark dataset}
\label{sec:observables}
To test the RL method introduced in this work, we construct a sufficiently complex benchmark environment. For this purpose, we consider a hypothetical dataset\footnote{Although real measurements are used for some observables, we refer to the overall dataset as hypothetical because it also contains fictitious measurements. In addition, the real measurements included in the benchmark are not necessarily up to date. The SM predictions and their corresponding uncertainties, on the other hand, are computed using \texttt{flavio}.} consisting of 132 observables drawn from several sectors of particle physics, including electroweak precision observables, $B$- and $K$-meson processes, $\tau$ decays, leptonic magnetic moments, Higgs observables, and scattering measurements. Within SMEFT, these sectors are not independent: correlations arise through gauge symmetry, RG evolution, matching, and dimension-six modifications of SM input parameters~\cite{Aebischer:2025qhh}.

Among the 132 observables in our dataset, 24 differ from their chosen SM predictions by more than $1.5\sigma$, resulting in $\chi^2_{\rm SM}=273$. These observables are listed in Tab.~\ref{tab:observables}. The remaining measurements act as constraints on the candidate operator sets.

The sole purpose of this dataset is to provide a challenging benchmark containing a complex pattern of discrepancies with respect to the SM for the RL search algorithm. Our aim is to test whether the RL policy can efficiently navigate a large operator space in the presence of simultaneous discrepancies across different physical processes, while accounting for the fact that operators that improve the description of one subset of observables may also affect many others.
\begin{table*}[h]
\centering
\small
\renewcommand{\arraystretch}{1.0}
\setlength{\tabcolsep}{5pt}
\captionsetup{width=1.0\textwidth}

\begin{tabular}{p{0.33\textwidth} c c c}
\hline
\rowcolor{gray!12}
Observable
& Pull
& Hypothetical measurement
& SM prediction \\
\hline

$\Delta M_s$
& $-5.06$
& $(8.10\pm0.49)\cdot 10^{-12}$
& $(1.14\pm0.046)\cdot 10^{-11}$ \\

$\epsilon'/\epsilon$
& $-4.53$
& $(-3.65\pm0.57)\cdot 10^{-3}$
& $(-2.76\pm55.57)\cdot 10^{-5}$ \\

$\mathrm{BR}(K^+\to\pi\nu\bar\nu)$
& $+4.32$
& $(1.12\pm0.051)\cdot 10^{-10}$
& $(8.14\pm0.50)\cdot 10^{-11}$ \\

$\langle P'_5\rangle(B^0\to K^*\mu\mu)_{[4,6]}$
& $-3.94$
& $-1.1607\pm0.0730$
& $-0.7583\pm0.0721$ \\

$m_W$
& $+3.83$
& $80.3992\pm0.0084$
& $80.3551\pm0.0078$ \\

$R_{\tau\ell}(B\to D\ell\nu)$
& $+3.79$
& $0.3275\pm0.0066$
& $0.2950\pm0.0056$ \\

$A_{\rm FB}(Z\to b\bar b)$
& $-3.38$
& $0.10239\pm0.00025$
& $0.10352\pm0.00022$ \\

$\langle d\mathrm{BR}/dq^2\rangle
(B_s\to\phi\mu\mu)_{[1,6]}$
& $-3.36$
& $(2.4556\pm0.6031)\cdot 10^{-8}$
& $(5.3556\pm0.6175)\cdot 10^{-8}$ \\

$\langle dR/d\theta\rangle(198.38,0.8,1.0)$
& $-2.97$
& $6.535\pm0.236$
& $7.236$ \\

$\mathrm{BR}(W\to\tau\nu)$
& $+2.56$
& $0.11380\pm0.00210$
& $0.108417\pm0.000049$ \\

$\mu_{tth}(h\to WW)$
& $+2.29$
& $5.00\pm1.75$
& $1.00$ \\

$\langle dR/d\theta\rangle(189.09,0.8,1.0)$
& $-2.20$
& $5.762\pm0.223$
& $6.253$ \\

$\mathrm{BR}(\tau\to\mu\nu\nu)$
& $+2.06$
& $0.17390\pm0.00040$
& $0.172894\pm0.000282$ \\

$\mu_{Zh}(h\to WW)$
& $+2.05$
& $5.90\pm2.39$
& $1.00$ \\

$\langle dR/d\theta\rangle(198.38,-0.6,-0.4)$
& $+1.86$
& $0.991\pm0.084$
& $0.835$ \\

$\langle dR/d\theta\rangle(198.38,0.6,0.8)$
& $-1.77$
& $4.100\pm0.185$
& $4.428$ \\

$\langle dR/d\theta\rangle(182.66,-1.0,-0.8)$
& $-1.75$
& $0.502\pm0.114$
& $0.702$ \\

$\epsilon_K$
& $+1.73$
& $(2.2280\pm0.0110)\cdot 10^{-3}$
& $(1.9529\pm0.1583)\cdot 10^{-3}$ \\

$\langle dR/d\theta\rangle(198.38,-1.0,-0.8)$
& $+1.73$
& $0.679\pm0.079$
& $0.542$ \\

$\langle dR/d\theta\rangle(182.66,0.0,0.2)$
& $+1.72$
& $2.150\pm0.244$
& $1.731$ \\

$\mu_{Wh}(h\to\tau\tau)$
& $-1.71$
& $-1.40\pm1.40$
& $1.00$ \\

$\langle dR/d\theta\rangle(205.92,-0.6,-0.4)$
& $-1.70$
& $0.653\pm0.069$
& $0.770$ \\

$R(e^+e^-\to WW)(182.7)$
& $+1.64$
& $1.036\pm0.022$
& $1.000$ \\

$\mu_{Zh}(h\to b\bar b)$
& $-1.50$
& $0.40\pm0.40$
& $1.00$ \\

\hline
\end{tabular}

\caption{
Anomalies (24 in total) included in the RL analysis, ordered by decreasing absolute
SM pull, $|p_i|$. The pull is defined as
$p_i=(O_i^{\rm exp}-O_i^{\rm SM})/
\sqrt{(\sigma_i^{\rm exp})^2+(\sigma_i^{\rm SM})^2}$.
The hypothetical experimental measurements and SM predictions
are quoted with their corresponding experimental and theoretical uncertainties,
respectively. The full dataset of 132 observables is available in the GitHub
repository \href{https://github.com/jackypheno/ReinforceSMEFT}{ReinforceSMEFT}.
}
\label{tab:observables}

\end{table*}

\section{Training of the RL Policy}
\label{sec:policy_training}
The objective of RL training is to learn a policy (or policies) through repeated interaction with 
the environment. The SMEFT operators that produce large improvements in the fit and satisfy the diversity criteria become more likely, whereas choices associated with poor trajectories are suppressed.

As an example, for training, we use 300 batches, with 25 stochastic trajectories sampled per
batch and at most $T=10$ proposals per trajectory. Since the RL search is sequential, in total, it  
involve $300\times 25\times 10=75$K calls to the internal $\chi^2$-minimization loop. Without 
using the pre-computed semi-analytical expressions, it would computationally very expensive.   
We use \textsc{Minuit}~\cite{James:1975dr} for the $\chi^2$ minimization. 

The actor and critic are optimized using AdamW with a learning rate of $10^{-3}$. We use a
discount factor $\gamma=0.99$, critic-loss coefficient
$\lambda_V=0.5$, and gradient clipping at $0.5$. An entropy term is
included during training to prevent premature collapse of the policy and
maintain exploration; its coefficient is annealed from $10^{-2}$ to
$10^{-4}$. The complete set of training hyperparameters and other details can be found on GitHub \href{https://github.com/jackypheno/ReinforceSMEFT}{ReinforceSMEFT}.

\begin{figure}[H]
    \centering
    \includegraphics[width=0.85\linewidth]
    {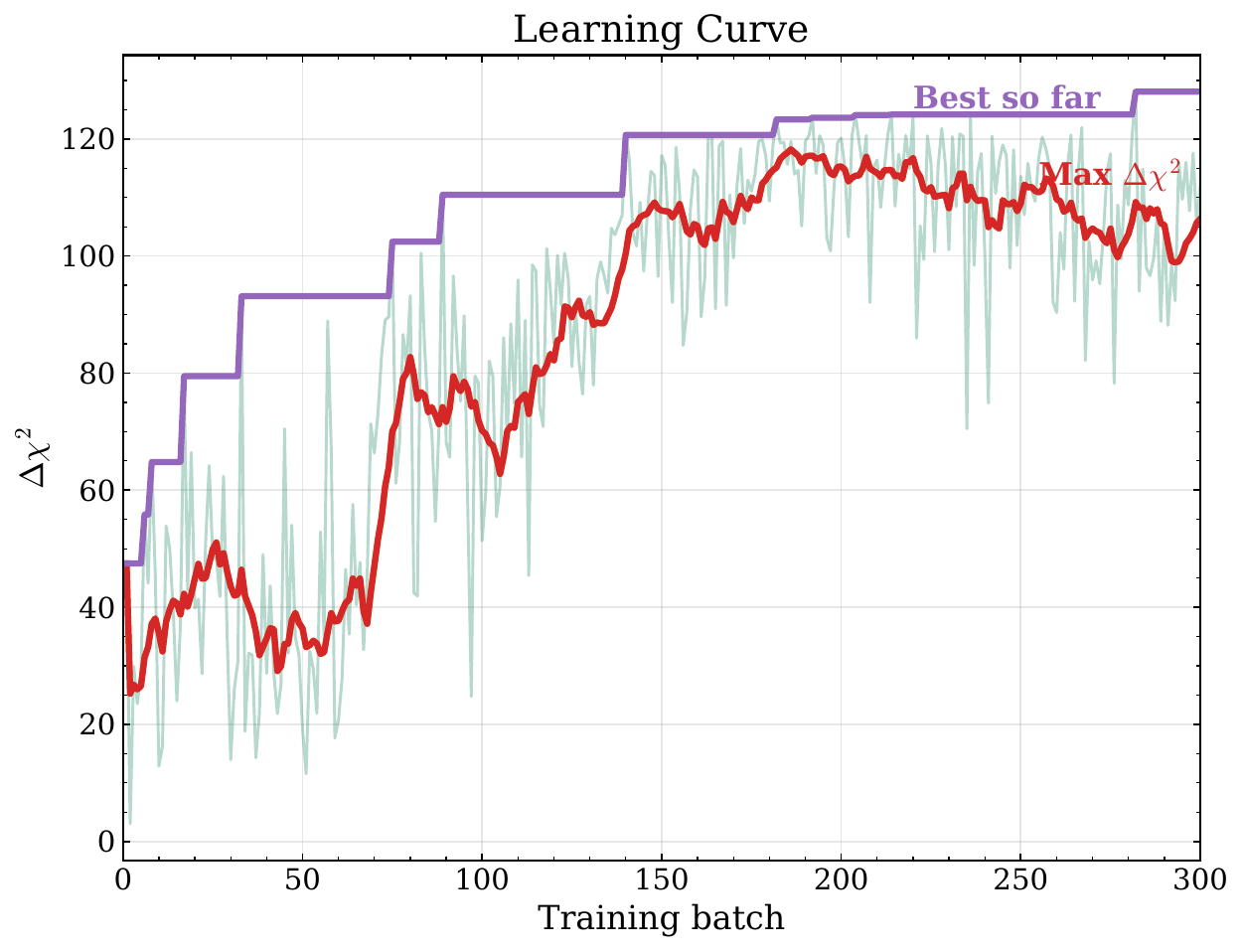}
    \caption{Learning curve of the RL policy for seed=42.
    For each training batch, the mean $\Delta\chi^2$
    relative to the SM is shown, together with the maximum 
    $\Delta\chi^2$ discovered up to that batch. The mean and maximum
    curves are smoothed using a 10-batch rolling average; the faint
    light green curve shows the corresponding unsmoothed values.}
    \label{fig:learning_curve}
\end{figure}

Fig.~\ref{fig:learning_curve} shows the evolution of fit quality
during training. For each batch, we display the mean and maximum
$\Delta\chi^2$ obtained by the sampled trajectories, together with the
best value discovered up to that point. The raw batch values fluctuate
substantially because the policy remains stochastic and is explicitly
encouraged to explore through both entropy regularization and diversity 
reward. Consequently, the mean or maximum $\Delta\chi^2$ of an
individual batch need not increase monotonically even after the policy
has discovered high-quality regions of the operator space.

The cumulative best curve provides a complementary view of learning:
once a high-quality solution is discovered, it remains part of a 
solution archive even while subsequent trajectories explore alternative
regions. The late-time fluctuations should therefore not be interpreted
as a loss of previously learned solutions, but rather as a consequence
of continued exploration in a quality--diversity search.

\section{Exploration of the SMEFT Operator Space}
\label{sec:solutions}

In this section, we analyze the viable SMEFT solutions\footnote{We use the terms solutions, hypotheses, theories or operator sets interchangeably throughout the text.  } discovered by 
the RL agent. To identify these solutions, post-training, 
we collect all candidate solutions sampled by the policy during 
training and rank them according to their $\Delta\chi^2$ values w.r.t. the SM.

We stress that, a policy optimized solely for fit quality can repeatedly return to the
neighborhood of a previously discovered high-quality operator set, generating
many solutions that differ by only one or a few operators. Such
solutions can be distinct as exact operator sets while representing
essentially the same underlying neighborhood in operator space. Our objective
is instead to discover both high-quality solutions and a broad range of
structurally distinct operator combinations.
The diversity term in Eq.~\eqref{eq:reward} addresses this issue by
rewarding solutions that have small Jaccard overlap with the existing
archive of high-quality theories. 

To quantify the resulting coverage of SMEFT 
hypotheses, we distinguish between \emph{exact solutions} -- corresponding to unique
operator sets, and \emph{solution families} -- which group sufficiently
similar operator sets according to their Jaccard similarity.

\paragraph{Exact solutions:}
Each candidate theory is specified by its set of active operators $S$. For illustration, 
we consider a theory a solution if it satisfies $\Delta\chi^2 > 100$ relative to
the SM, and we count two solutions as the same \emph{exact solution} if they
contain identical operator sets, irrespective of operator order.

\paragraph{Solution families:}
 To
quantify how many structurally different solutions are found, we additionally
group exact solutions into \emph{families} using the Jaccard similarity.
Solutions are processed in the order in which they are discovered, batch by
batch. A new solution $S$ is compared with the stored family representatives
$\{R_k\}$. If $J(S,R_k) > 0.5$ for any $k$, $S$ is assigned to an existing
family and the number of families is unchanged. Otherwise $S$ becomes the
\emph{founder} of a new family and is stored as its representative. Thus the
founder of a family is the first solution encountered that is not similar
to any earlier representative, and is not necessarily the solution with the
largest $\Delta\chi^2$. 

Because membership is defined only relative to the representative,
it is not transitive, and the precise family count depends mildly on the
discovery order. The same procedure and threshold are applied to the RL
archive and to both random baselines as discussed below.

\subsection{Comparison with random-search baselines}

\begin{figure}[H]
    \centering
    \includegraphics[width=0.85\linewidth]
    {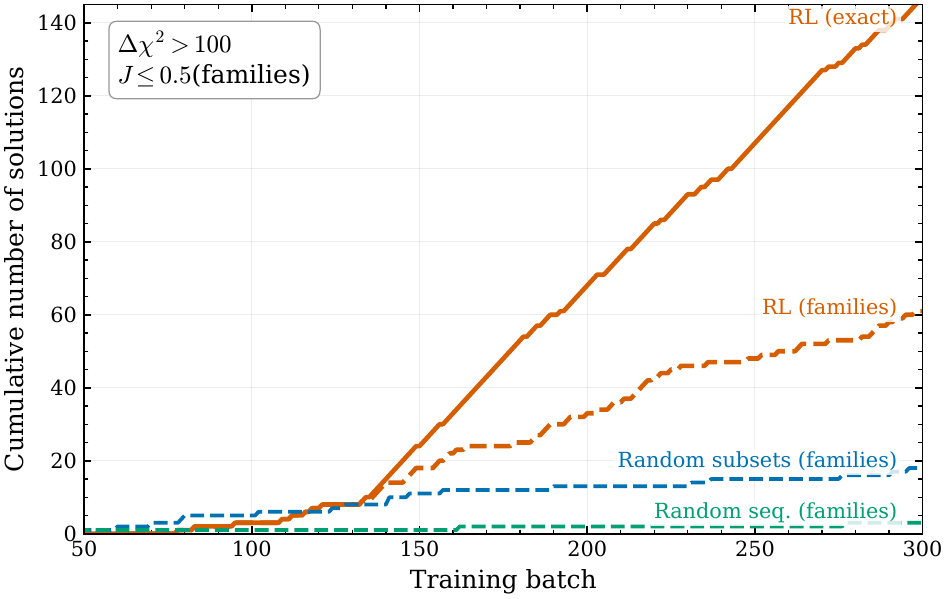}
    \caption{Cumulative number of high-quality SMEFT solutions
    discovered during the training as a function of batch. Only theories with
    $\Delta\chi^2>100$ relative to the SM are retained. Exact operator
    sets and Jaccard families are shown for the RL
    agent and the random-search baselines. The random-sequential
    baseline follows the same step-wise construction and acceptance
    procedure as the RL agent but proposes operators uniformly at
    random.}
    \label{fig:rl_vs_random}
\end{figure}

To benchmark the RL search, we consider two random-search baselines with comparable computational budgets. In the \emph{random-subset} baseline, the number of operators in each subset is chosen randomly, and the corresponding operators are then sampled independently from the full operator space and evaluated directly. This provides a simple measure of how frequently high-quality hypotheses can be found through unguided sampling of the combinatorial operator space.

The second benchmark, the \emph{random-sequential} baseline, follows the
same sequential construction and acceptance rule as the RL agent, but
selects each proposed operator uniformly at random. This provides a
more direct algorithmic comparison because both methods construct theories
sequentially and receive the same physics evaluation, but only the RL
agent uses a learned state-dependent policy to select the next operator.

Fig.~\ref{fig:rl_vs_random} compares the cumulative discovery of
solutions satisfying $\Delta\chi^2>100$ during the RL training. This threshold deliberately
focuses the comparison on high-quality theories rather than on the total
number of sampled configurations. The exact-solution count measures the
number of unique operator sets discovered, whereas the family count
provides a more stringent measure of operator diversity defined by $J\le 0.5$.
The comparison therefore probes two complementary aspects of the search: (i)
whether the learned policy can concentrate on SMEFT operator space 
 containing good fits and (ii) whether it can continue to discover
distinct regions rather than repeatedly sampling small variations of
the same operator set, e.g. by changing one SMEFT operator. 

These results show that the RL search fulfills our objective of not only 
maximizing $\Delta\chi^2$ but also efficiently exploring the high-quality 
regions of the SMEFT operator space.

\subsection{Coverage across independent training seeds}
Because RL training is stochastic, independent runs can converge toward
different regions of the SMEFT operator space. However, this feature can be
exploited by combining the solution archives obtained from multiple
training seeds.
\begin{figure}[h]
    \centering
    \includegraphics[width=0.85\linewidth]
    {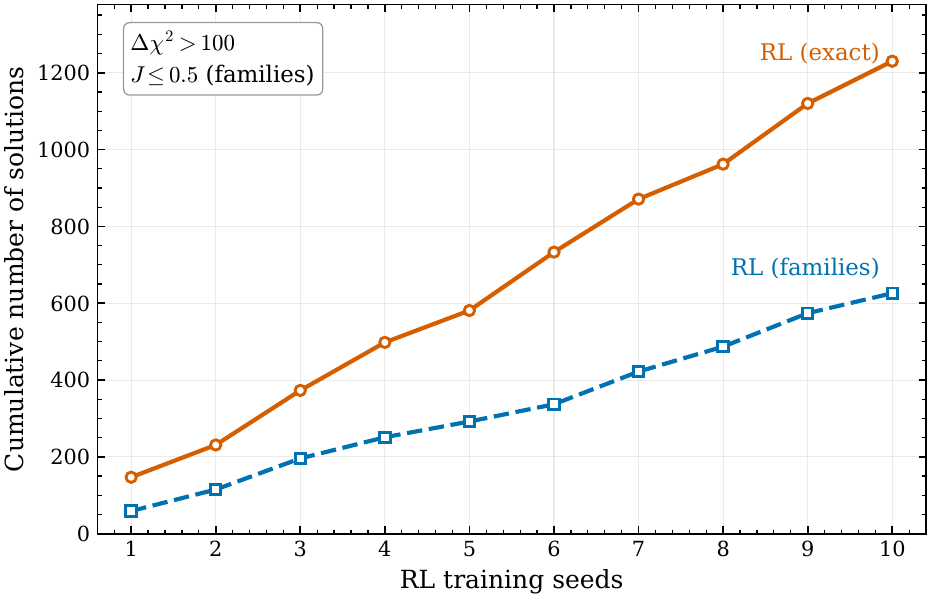}
    \caption{Cumulative coverage of the SMEFT solution space obtained
    by combining independent RL training runs. Only theories satisfying
    $\Delta\chi^2>100$ are included. The curves show the cumulative
    numbers of distinct exact operator sets and Jaccard 
    families as additional training seeds are included.}
    \label{fig:rl_seed_coverage}
\end{figure}
Fig.~\ref{fig:rl_seed_coverage} shows the cumulative coverage obtained
by combining ten independently trained RL policies. Exact operator sets
typically grow more rapidly than the number of families because
different runs can discover variations of the same underlying
structures. The family curve therefore provides the more informative
measure of whether additional independent training runs continue to
probe genuinely different regions of operator space.

In this figure, the continued
growth of the family count indicates that additional runs still uncover
new structures, whereas saturation would suggest that the dominant
families accessible to the search have largely been identified.
In total, we identified 26 solutions with $\Delta \chi^2>100$ and with $J \le 0.1$,  
which are listed in Tab.~\ref{tab:distinct_rl_solutions} along with corresponding $\Delta \chi^2$ 
values w.r.t. the SM hypothesis.  The corresponding definitions of the SMEFT operators 
are provided in Tab.~\ref{tab:warsaw_operators_rl}.
\begin{table*}[t]
\centering
\small
\renewcommand{\arraystretch}{1.18}
\setlength{\tabcolsep}{5pt}

\begin{tabular}{c p{0.7\textwidth} c}
\hline
\rowcolor{gray!12}
\# & WCs & $\Delta\chi^2$ \\

\hline

1 &
$\wc[(1)]{lequ}{3332}$,
$\wc[(3)]{lequ}{2232}$,
$\wc[]{Hd}{33}$,
$\wc[(3)]{Hq}{12}$,
$\wc[]{Hu}{33}$,
$\wc[(1)]{quqd}{2311}$,
$\wc[(1)]{quqd}{2323}$
& 142.9 \\

2 &
$\wc[]{dB}{33}$,
$\wc[]{dG}{22}$,
$\wc[]{ld}{1112}$,
$\wc[]{ledq}{3333}$,
$\wc[(3)]{lequ}{2213}$,
$\wc[]{Hd}{33}$,
$\wc[(1)]{qd}{2223}$,
$\wc[(8)]{qd}{1321}$,
$\wc[(8)]{qd}{2223}$,
$\wc[(1)]{quqd}{1321}$
& 137.6 \\

3 &
$\wc[]{dW}{22}$,
$\wc[]{eW}{11}$,
$\wc[]{ld}{1123}$,
$\wc[]{ledq}{3332}$,
$\wc[(1)]{lq}{1123}$,
$\wc[]{Hd}{33}$,
$\wc[(1)]{qq}{2223}$,
$\wc[(8)]{quqd}{3312}$,
$\wc[(1)]{ud}{2321}$
& 135.7 \\

4 &
$\wc[]{dG}{22}$,
$\wc[]{dd}{2223}$,
$\wc[]{ed}{3312}$,
$\wc[]{le}{1221}$,
$\wc[(3)]{lequ}{2232}$,
$\wc[]{ll}{1221}$,
$\wc[(3)]{lq}{1123}$,
$\wc[(1)]{qd}{1321}$,
$\wc[(8)]{qu}{2333}$,
$\wc[(1)]{quqd}{3321}$
& 130.0 \\

5 &
$\wc[]{dW}{21}$,
$\wc[]{eW}{22}$,
$\wc[]{eu}{1111}$,
$\wc[]{ledq}{3333}$,
$\wc[(1)]{Hq}{13}$,
$\wc[(1)]{qd}{1113}$,
$\wc[(1)]{qd}{3333}$,
$\wc[(1)]{quqd}{2323}$
& 129.5 \\

6 &
$\wc[]{dG}{12}$,
$\wc[(3)]{lequ}{2233}$,
$\wc[(3)]{lequ}{3332}$,
$\wc[(1)]{lq}{2212}$,
$\wc[(8)]{qd}{1213}$,
$\wc[(1)]{quqd}{2323}$,
$\wc[(1)]{ud}{3333}$
& 127.0 \\

7 &
$\wc[]{dW}{21}$,
$\wc[]{eB}{22}$,
$\wc[]{ld}{1123}$,
$\wc[(1)]{lequ}{3332}$,
$\wc[]{ll}{2332}$,
$\wc[(3)]{lq}{1123}$,
$\wc[(1)]{qd}{2312}$,
$\wc[(1)]{qu}{2323}$
& 125.5 \\

8 &
$\wc[]{ld}{3323}$,
$\wc[]{ledq}{3333}$,
$\wc[(3)]{lequ}{2223}$,
$\wc[(3)]{lq}{2233}$,
$\wc[(1)]{qd}{1221}$,
$\wc[(8)]{qd}{2333}$,
$\wc[(3)]{qq}{1213}$,
$\wc[(8)]{quqd}{3121}$,
$\wc[(8)]{quqd}{3222}$
& 124.4 \\

9 &
$\wc[]{dB}{12}$,
$\wc[]{eW}{11}$,
$\wc[(1)]{lq}{3323}$,
$\wc[(1)]{Hl}{11}$,
$\wc[(3)]{qq}{1123}$,
$\wc[(3)]{qq}{1233}$,
$\wc[(8)]{qu}{2333}$,
$\wc[(1)]{ud}{1123}$
& 119.6 \\

10 &
$\wc[]{W}{}$,
$\wc[]{ed}{1123}$,
$\wc[(3)]{lequ}{2213}$,
$\wc[(3)]{lq}{1123}$,
$\wc[(1)]{Hq}{13}$,
$\wc[(1)]{qq}{2223}$,
$\wc[(8)]{quqd}{3121}$
& 118.3 \\

11 &
$\wc[(3)]{lequ}{2213}$,
$\wc[(3)]{lequ}{2232}$,
$\wc[(1)]{lq}{1112}$,
$\wc[(1)]{Hq}{22}$,
$\wc[(1)]{qd}{1221}$,
$\wc[(1)]{qu}{2323}$,
$\wc[(1)]{quqd}{3222}$,
$\wc[(1)]{quqd}{3332}$,
$\wc[(8)]{quqd}{3112}$
& 114.6 \\

12 &
$\wc[]{dG}{22}$,
$\wc[(1)]{lequ}{3332}$,
$\wc[(3)]{lequ}{2233}$,
$\wc[(8)]{qd}{2333}$,
$\wc[(1)]{qq}{1123}$,
$\wc[(3)]{qq}{1223}$
& 113.2 \\

13 &
$\wc[]{dd}{2333}$,
$\wc[]{ld}{1112}$,
$\wc[(1)]{lequ}{3332}$,
$\wc[(1)]{Hl}{11}$,
$\wc[(8)]{qu}{1311}$,
$\wc[(8)]{quqd}{2332}$,
$\wc[(8)]{ud}{1112}$
& 111.9 \\

14 &
$\wc[]{dW}{22}$,
$\wc[]{eB}{22}$,
$\wc[]{lu}{1111}$,
$\wc[(1)]{qd}{1321}$,
$\wc[(1)]{quqd}{2323}$,
$\wc[(1)]{ud}{2223}$
& 109.7 \\

15 &
$\wc[(1)]{lequ}{1133}$,
$\wc[(3)]{lq}{2212}$,
$\wc[(1)]{qd}{1123}$,
$\wc[(1)]{qd}{1213}$,
$\wc[(8)]{qd}{2333}$,
$\wc[(1)]{qq}{2223}$,
$\wc[(3)]{qq}{2222}$,
$\wc[(1)]{quqd}{3111}$,
$\wc[(8)]{ud}{3312}$
& 109.2 \\

16 &
$\wc[]{dG}{12}$,
$\wc[]{dW}{31}$,
$\wc[(3)]{lequ}{2212}$,
$\wc[(1)]{lq}{1113}$,
$\wc[(3)]{lq}{3312}$,
$\wc[(8)]{qd}{3323}$,
$\wc[(1)]{qu}{2323}$,
$\wc[(8)]{quqd}{3312}$
& 108.7 \\

17 &
$\wc[]{dB}{31}$,
$\wc[]{eW}{22}$,
$\wc[(3)]{lq}{1123}$,
$\wc[(8)]{qu}{2323}$,
$\wc[(1)]{quqd}{2321}$,
$\wc[(8)]{quqd}{2311}$
& 107.3 \\

18 &
$\wc[]{dd}{2333}$,
$\wc[]{ed}{1122}$,
$\wc[(3)]{lq}{1112}$,
$\wc[(3)]{lq}{1123}$,
$\wc[(1)]{qd}{1113}$,
$\wc[(1)]{quqd}{2131}$
& 106.0 \\

19 &
$\wc[]{dB}{11}$,
$\wc[]{dG}{12}$,
$\wc[]{dG}{23}$,
$\wc[(1)]{lequ}{2233}$,
$\wc[(1)]{lq}{3312}$,
$\wc[(8)]{qd}{2333}$,
$\wc[(1)]{quqd}{2321}$
& 106.0 \\

20 &
$\wc[]{dW}{22}$,
$\wc[]{dd}{2323}$,
$\wc[]{eW}{22}$,
$\wc[(1)]{lq}{1113}$,
$\wc[(3)]{qq}{1123}$,
$\wc[(3)]{qq}{1332}$,
$\wc[(1)]{quqd}{3211}$,
$\wc[(1)]{quqd}{3222}$
& 105.7 \\

21 &
$\wc[(1)]{lequ}{2233}$,
$\wc[(8)]{qd}{2312}$,
$\wc[(8)]{qd}{2323}$,
$\wc[(3)]{qq}{1123}$,
$\wc[(3)]{qq}{1322}$,
$\wc[(8)]{quqd}{3312}$
& 105.0 \\

22 &
$\wc[]{dG}{22}$,
$\wc[(1)]{lequ}{1133}$,
$\wc[(3)]{lq}{1112}$,
$\wc[(8)]{qu}{1311}$,
$\wc[(1)]{ud}{2323}$,
$\wc[(8)]{ud}{1321}$
& 104.0 \\

23 &
$\wc[]{Hd}{33}$,
$\wc[(8)]{qd}{1211}$,
$\wc[(8)]{qd}{2323}$,
$\wc[(8)]{qd}{3312}$,
$\wc[(1)]{quqd}{3112}$,
$\wc[(8)]{quqd}{3222}$
& 103.0 \\

24 &
$\wc[]{eW}{11}$,
$\wc[(1)]{qd}{1221}$,
$\wc[(1)]{qd}{2332}$,
$\wc[(8)]{qd}{3323}$,
$\wc[(1)]{quqd}{3323}$,
$\wc[(8)]{ud}{1321}$
& 101.0 \\

25 &
$\wc[]{dd}{2333}$,
$\wc[]{ld}{1133}$,
$\wc[]{ledq}{3332}$,
$\wc[]{qe}{1233}$,
$\wc[(8)]{quqd}{1132}$,
$\wc[(8)]{quqd}{3323}$
& 100.5 \\

26 &
$\wc[(3)]{lequ}{2222}$,
$\wc[(3)]{lq}{1111}$,
$\wc[(8)]{qd}{1112}$,
$\wc[(1)]{qq}{1223}$,
$\wc[(1)]{quqd}{2323}$,
$\wc[(1)]{quqd}{3323}$,
$\wc[(8)]{quqd}{3111}$
& 100.3 \\

\hline
\end{tabular}

\caption{
Distinct SMEFT solution families ($J \le 0.1$) identified by the RL agent in 
ten different training runs with different seeds. The solutions are ordered by decreasing
$\Delta\chi^2$, and here $\chi^2_{\rm SM}=273$. The 26 solutions contain 122 distinct flavor-specific
WCs. We remind that the total number of observables included in the fit are 132. The the definitions of 
corresponding SMEFT operators are given in Tab.~\ref{tab:warsaw_operators_rl}.
}
\label{tab:distinct_rl_solutions}
\end{table*}

% ============================================================
% CONCLUSION
% ============================================================

\section{Conclusions and Future Prospects}
\label{sec:conclusions}

In order to interpret complex patterns of anomalies with respect to the SM, we have formulated an RL method that can be used to search for viable SMEFT operator hypotheses sequentially. This addresses the central challenge of the combinatorial explosion of possible hypotheses within SMEFT. Our formulation provides a novel approach to this problem by learning an ``actor--critic'' RL policy.

Importantly, our objective differs from a naive optimization problem in an important respect. We are not interested only in the SMEFT hypothesis that maximizes the improvement in $\chi^2$. Because several structurally different SMEFT scenarios may provide good descriptions of the same data, mapping this degeneracy is crucial for NP model building in a bottom-up approach. We therefore introduced a diversity reward based on Jaccard similarity, turning the search into a quality--diversity problem in which the RL policy is rewarded both for improving the global fit and for exploring operator sets different from those already discovered.

We tested the framework in a challenging environment containing 132 observables from several sectors, including 24 hypothetical measurements with discrepancies larger than $1.5\sigma$ relative to the SM predictions. The RL agent searched over 912 candidate SMEFT operators and constructed hypotheses sequentially while accounting for the effects of each operator proposal on all observables. The resulting learning behaviour shows that the policy is able to identify 26 high-quality solutions with Jaccard indices below 0.1. Our comparisons with random-subset and random-sequential search baselines provide a direct benchmark of both search efficiency and solution diversity, while independent training seeds offer a systematic way of extending the coverage of the operator space.

Several extensions are natural. The present framework can be applied to larger observable sets and updated experimental measurements, and the quality--diversity objective can be refined to target specific requirements of the physics problem. It will also be useful to investigate how policies trained on one pattern of experimental discrepancies transfer to modified datasets, and whether the learned search strategy can be amortized across related SMEFT inference problems.

More broadly, the approach developed here illustrates how RL can complement conventional phenomenological reasoning by systematically exploring theory spaces whose combinatorial complexity makes exhaustive human-guided analysis increasingly difficult. To facilitate the broader application of our framework, we also provide an open-source Python package, \href{https://github.com/jackypheno/ReinforceSMEFT}{ReinforceSMEFT}, that can be readily installed via \texttt{pip}, enabling users to train and apply the RL agent to their own phenomenological analyses.
\section*{Acknowledgments}
We gratefully acknowledge OpenAI for providing academic access to ChatGPT, which supported aspects of this research. This work was financially supported by the Natural Sciences and Engineering Research Council of Canada (NSERC). This research was enabled in part by support provided by the Digital Research Alliance of Canada (\url{https://alliancecan.ca}). Computations were performed on the Trillium supercomputer at the SciNet \cite{Loken_2010} HPC Consortium. SciNet is funded by Innovation, Science and Economic Development Canada; the Digital Research Alliance of Canada; the Ontario Research Fund: Research Excellence; and the University of Toronto.

\section*{Appendix}

In Tab.~\ref{tab:warsaw_operators_rl}, we provide the definitions of the Warsaw-basis operators that appear in the solutions identified by the RL search. A complete list of Warsaw-basis operators can be found in Refs.~\cite{Grzadkowski:2010es,Aebischer:2017ugx}. Throughout this work, we adopt the SMEFT down-basis flavor convention. The complete list of operators, including their flavor indices in the WCxf convention, is available on the \href{https://wcxf.github.io/assets/pdf/SMEFT.Warsaw.pdf}{WCxf Warsaw basis} webpage.

\begin{table*}[h]
\centering
\scriptsize
\renewcommand{\arraystretch}{1.18}
\setlength{\tabcolsep}{5pt}

\begin{tabularx}{\textwidth}{
    >{\centering\arraybackslash}p{0.10\textwidth}
    >{\raggedright\arraybackslash}X
    >{\centering\arraybackslash}p{0.10\textwidth}
    >{\raggedright\arraybackslash}X
}
\toprule

\multicolumn{2}{c}{
\cellcolor{gray!12}
\textbf{Bosonic and two-fermion operators}
}
&
\multicolumn{2}{c}{
\cellcolor{gray!12}
\textbf{Higgs--fermion current operators}
}
\\
\midrule

$\mathcal O_W$
&
$\epsilon^{IJK}
W_{\mu}^{I\nu}
W_{\nu}^{J\rho}
W_{\rho}^{K\mu}$
&
$\mathcal O_{Hq}^{(1)}$
&
$(H^\dagger i\overleftrightarrow D_\mu H)
(\bar q_p\gamma^\mu q_r)$
\\

$\mathcal O_{dG}$
&
$(\bar q_p\sigma^{\mu\nu}T^A d_r)H\,G_{\mu\nu}^{A}$
&
$\mathcal O_{Hq}^{(3)}$
&
$(H^\dagger i\overleftrightarrow D_\mu^{\,I}H)
(\bar q_p\tau^I\gamma^\mu q_r)$
\\

$\mathcal O_{dW}$
&
$(\bar q_p\sigma^{\mu\nu}d_r)\tau^I H\,W_{\mu\nu}^{I}$
&
$\mathcal O_{Hl}^{(1)}$
&
$(H^\dagger i\overleftrightarrow D_\mu H)
(\bar l_p\gamma^\mu l_r)$
\\

$\mathcal O_{dB}$
&
$(\bar q_p\sigma^{\mu\nu}d_r)H\,B_{\mu\nu}$
&
$\mathcal O_{Hu}$
&
$(H^\dagger i\overleftrightarrow D_\mu H)
(\bar u_p\gamma^\mu u_r)$
\\

$\mathcal O_{eW}$
&
$(\bar l_p\sigma^{\mu\nu}e_r)\tau^I H\,W_{\mu\nu}^{I}$
&
$\mathcal O_{Hd}$
&
$(H^\dagger i\overleftrightarrow D_\mu H)
(\bar d_p\gamma^\mu d_r)$
\\

$\mathcal O_{eB}$
&
$(\bar l_p\sigma^{\mu\nu}e_r)H\,B_{\mu\nu}$
&
&
\\

\midrule

\multicolumn{2}{c}{
\cellcolor{gray!12}
\textbf{$(\bar L L)(\bar L L)$}
}
&
\multicolumn{2}{c}{
\cellcolor{gray!12}
\textbf{$(\bar R R)(\bar R R)$}
}
\\
\midrule

$\mathcal O_{ll}$
&
$(\bar l_p\gamma_\mu l_r)
(\bar l_s\gamma^\mu l_t)$
&
$\mathcal O_{dd}$
&
$(\bar d_p\gamma_\mu d_r)
(\bar d_s\gamma^\mu d_t)$
\\

$\mathcal O_{qq}^{(1)}$
&
$(\bar q_p\gamma_\mu q_r)
(\bar q_s\gamma^\mu q_t)$
&
$\mathcal O_{eu}$
&
$(\bar e_p\gamma_\mu e_r)
(\bar u_s\gamma^\mu u_t)$
\\

$\mathcal O_{qq}^{(3)}$
&
$(\bar q_p\gamma_\mu\tau^I q_r)
(\bar q_s\gamma^\mu\tau^I q_t)$
&
$\mathcal O_{ed}$
&
$(\bar e_p\gamma_\mu e_r)
(\bar d_s\gamma^\mu d_t)$
\\

$\mathcal O_{lq}^{(1)}$
&
$(\bar l_p\gamma_\mu l_r)
(\bar q_s\gamma^\mu q_t)$
&
$\mathcal O_{ud}^{(1)}$
&
$(\bar u_p\gamma_\mu u_r)
(\bar d_s\gamma^\mu d_t)$
\\

$\mathcal O_{lq}^{(3)}$
&
$(\bar l_p\gamma_\mu\tau^I l_r)
(\bar q_s\gamma^\mu\tau^I q_t)$
&
$\mathcal O_{ud}^{(8)}$
&
$(\bar u_p\gamma_\mu T^A u_r)
(\bar d_s\gamma^\mu T^A d_t)$
\\

\midrule

\multicolumn{2}{c}{
\cellcolor{gray!12}
\textbf{$(\bar L L)(\bar R R)$}
}
&
\multicolumn{2}{c}{
\cellcolor{gray!12}
\textbf{Scalar and tensor structures}
}
\\
\midrule

$\mathcal O_{le}$
&
$(\bar l_p\gamma_\mu l_r)
(\bar e_s\gamma^\mu e_t)$
&
$\mathcal O_{ledq}$
&
$(\bar l_p^{\,j}e_r)
(\bar d_s q_t^{\,j})$
\\

$\mathcal O_{lu}$
&
$(\bar l_p\gamma_\mu l_r)
(\bar u_s\gamma^\mu u_t)$
&
$\mathcal O_{lequ}^{(1)}$
&
$(\bar l_p^{\,j}e_r)
\epsilon_{jk}
(\bar q_s^{\,k}u_t)$
\\

$\mathcal O_{ld}$
&
$(\bar l_p\gamma_\mu l_r)
(\bar d_s\gamma^\mu d_t)$
&
$\mathcal O_{lequ}^{(3)}$
&
$(\bar l_p^{\,j}\sigma_{\mu\nu}e_r)
\epsilon_{jk}
(\bar q_s^{\,k}\sigma^{\mu\nu}u_t)$
\\

$\mathcal O_{qe}$
&
$(\bar q_p\gamma_\mu q_r)
(\bar e_s\gamma^\mu e_t)$
&
$\mathcal O_{quqd}^{(1)}$
&
$(\bar q_p^{\,j}u_r)
\epsilon_{jk}
(\bar q_s^{\,k}d_t)$
\\

$\mathcal O_{qu}^{(1)}$
&
$(\bar q_p\gamma_\mu q_r)
(\bar u_s\gamma^\mu u_t)$
&
$\mathcal O_{quqd}^{(8)}$
&
$(\bar q_p^{\,j}T^A u_r)
\epsilon_{jk}
(\bar q_s^{\,k}T^A d_t)$
\\

$\mathcal O_{qu}^{(8)}$
&
$(\bar q_p\gamma_\mu T^A q_r)
(\bar u_s\gamma^\mu T^A u_t)$
&
&
\\

$\mathcal O_{qd}^{(1)}$
&
$(\bar q_p\gamma_\mu q_r)
(\bar d_s\gamma^\mu d_t)$
&
&
\\

$\mathcal O_{qd}^{(8)}$
&
$(\bar q_p\gamma_\mu T^A q_r)
(\bar d_s\gamma^\mu T^A d_t)$
&
&
\\

\bottomrule
\end{tabularx}

\caption{
Warsaw-basis \cite{Grzadkowski:2010es} dimension-six operators appearing in the RL solutions.
The indices $p,r,s,t$ denote fermion generations, while $j,k$ are
$SU(2)_L$ indices. The matrices $\tau^I$ and $T^A$ denote the
$SU(2)_L$ and $SU(3)_c$ generators, respectively, and
$\epsilon_{jk}$ is the antisymmetric $SU(2)_L$ tensor.
}
\label{tab:warsaw_operators_rl}
\end{table*}

\newpage

\bibliographystyle{plainnat}
\bibliography{NP_with_RL}
\end{document}